\pdfoutput=1

\documentclass[11pt]{article}

\usepackage[]{acl}

\usepackage{times}
\usepackage{latexsym}

\usepackage[T1]{fontenc}

\usepackage[utf8]{inputenc}

\usepackage{microtype}

\usepackage{float}
\usepackage{enumitem}
\setenumerate[1]{itemsep=0pt,partopsep=0pt,parsep=\parskip,topsep=0pt}
\setitemize[1]{itemsep=0pt,partopsep=0pt,parsep=\parskip,topsep=0pt}
\setdescription{itemsep=0pt,partopsep=0pt,parsep=\parskip,topsep=0pt}

\usepackage{CJKutf8}
\usepackage{graphicx}
\usepackage{bm}
\usepackage{amsmath}
\usepackage{multirow,booktabs}
\usepackage{ulem}
\usepackage{adjustbox}

\definecolor{CadmiumOrange}{rgb}{0.93,0.53, 0.18}
\definecolor{ForestGreen}{rgb}{0.13, 0.55, 0.13}

\title{CLOWER: A Pre-trained Language Model with Contrastive Learning over Word and Character Representations}

\author{Borun Chen\textsuperscript{1}\thanks{\enspace Equal contribution.}, Hongyin Tang\textsuperscript{2}\footnotemark[1], Jiahao Bu\textsuperscript{2}, Kai Zhang\textsuperscript{2}, Jingang Wang\textsuperscript{2}\thanks{\enspace Corresponding authors.},\\ \textbf{Qifan Wang\textsuperscript{3},
  Hai-Tao Zheng\textsuperscript{1, 4}\footnotemark[2], Wei Wu\textsuperscript{2} and Liqian Yu\textsuperscript{2} }\\
  \textsuperscript{1}Tsinghua Shenzhen International Graduate School, Tsinghua University \\
  \textsuperscript{2}Meituan\quad\textsuperscript{3}Meta AI\quad\textsuperscript{4}Peng Cheng Laboratory \\
  \texttt{cbr20@mails.tsinghua.edu.cn zheng.haitao@sz.tsinghua.edu.cn} \\
  \texttt{\{tanghongyin,wangjingang02,wuwei30,yuliqian\}@meituan.com}\\
  \texttt{bujh1994@gmail.com zhangkai.plus@foxmail.com wqfcr@fb.com}}
  
\newcommand{\M}{CLOWER}

\begin{document}

\maketitle
\begin{abstract}
Pre-trained Language Models (PLMs) have achieved remarkable performance gains across numerous downstream tasks in natural language understanding. Various Chinese PLMs have been successively proposed for learning better Chinese language representation.
However, most current models use Chinese characters as inputs and are not able to encode semantic information contained in Chinese words.
While recent pre-trained models incorporate both words and characters simultaneously, they usually suffer from deficient semantic interactions and fail to capture the semantic relation between words and characters.
To address the above issues, we propose a simple yet effective PLM CLOWER, which adopts the Contrastive Learning Over Word and charactER representations. 
In particular, CLOWER implicitly encodes the coarse-grained information (i.e., words) into the fine-grained representations (i.e., characters) through contrastive learning on multi-grained information.
CLOWER is of great value in realistic scenarios since it can be easily incorporated into any existing fine-grained based PLMs without modifying the production pipelines.
Extensive experiments conducted on a range of downstream tasks demonstrate the superior performance of CLOWER over several state-of-the-art baselines.
\end{abstract}

\section{Introduction}
Pre-trained language models (PLMs) have gained tremendous success in the field of natural language processing recently. 
As a major milestone of PLMs, BERT~\citep{kenton2019bert} and its variants~\citep{yang2019xlnet, liu2019roberta, clark2019electra} have demonstrated outstanding performance on a wide variety of natural language understanding (NLU) tasks, such as sentiment analysis and machine reading comprehension tasks. The architecture of Transformer~\citep{vaswani2017attention} is typically the foundation for these models, which models the semantic and syntactic relationships between the tokens of the entire input text and learns the contextual representations for each token.

Early Chinese PLMs~\citep{sun2019ernie} often take the sequences of Chinese characters as the input. 
These models require relatively small vocabulary and learn the representations of each character from the corpus, which avoids the Out-Of-Vocabulary problem~\citep{li2019word}. 
However, the meanings of a Chinese word can be totally different from the meanings of each Chinese character in the word. 
For example, the meaning of \begin{CJK*}{UTF8}{gbsn}“小心”\end{CJK*} (careful) can not be derived from summing the meaning of \begin{CJK*}{UTF8}{gbsn}“小”\end{CJK*} (small) and \begin{CJK*}{UTF8}{gbsn}“心”\end{CJK*} (heart). 
In general, the phenomenon of semantic gaps between coarse-grained language units and fine-grained language units (e.g., words \& characters, phrases \& words) exists not only in Chinese but also in many other languages.

To alleviate the gap, prior studies improve the pre-trained models in two directions. 
One direction is to enrich the masking strategies in the masked language model (MLM) objective to mask coarse-grained units, such as the whole word masking (WWM)~\citep{cui2021pre} and phrase masking~\citep{sun2019ernie}. 
These methods encourage the pre-trained model to recover the coarse-grained masks with fine-grained tokens. 
However, the relation between the coarse-grained and fine-grained representations is modeled in an implicit manner, leading to less effective representations.
The other direction is to leverage the multi-grained tokenizations as input.
AMBERT~\citep{zhang2021ambert} encodes both the fine-grained and coarse-grained token sequences and performs the masked language modeling tasks correspondingly, while LICHEE~\citep{guo2021lichee} merges the multi-grained token embeddings explicitly to integrate the information. 
Lattice-BERT~\citep{lai2021lattice} adopts the lattice graph to construct the multi-grained input. 
Nevertheless, these models require additional computational costs (e.g., tokenization, graph construction, multi-grained encoding) and the multi-grained information is only integrated in the embedding layer other than the full encoder, leading to limited usability with low effectiveness.

To fully leverage the semantic information of multi-granularity and preserve the flexibility of single-grained models in the fine-tuning stage, we propose a novel PLM named \M~to efficiently model the multi-grained semantic information in pre-training to improve the representation capability. 
\M~adopts the contrastive learning framework to carry out the semantic interaction between multi-grained representations. 
Specifically, in the pre-training stage, we perform both character and word level tokenizations separately for each input sequence and feed them into the encoder to obtain the contextual representations.
Then we conduct the contrastive learning over character and word representations on both token-level and sentence-level. 
In this way, the word-level semantic information is encoded into the character tokens by bringing their representations closer.
Different from AMBERT or LICHEE, in fine-tuning, \M~requires no additional computation and can be directly used in any fine-grained PLMs.
The merit makes CLOWER production-friendly since it could be deployed easily without modifying the established production pipeline.

We perform comprehensive experiments on different downstream NLU tasks. 
The experimental results show that \M~achieves considerable improvements over several baselines.
Ablation studies demonstrate the effectiveness of contrastive learning in our pre-training framework.
Our contributions are summarized as follows:
\begin{itemize}
    \item We present a novel approach that adopts contrastive learning over both word and character representations, which effectively captures  their semantic relations.
    \item With the help of the aforementioned contrastive learning approach, we introduce a Chinese pre-trained language model that connects multi-grained semantic information for learning high quality word and character encoders. 
    \item We conduct an extensive set of experiments on several benchmarks and demonstrate the effectiveness of the proposed model.
\end{itemize}

\section{Related Work}
\paragraph{Multi-grained Pre-trained Language Models} There have been some efforts to explore the multi-granularity information on the pre-trained language models~\citep{tay2021charformer, xue2022byt5}. \citet{cui2021pre} adopts the whole word masking strategy to select the masking tokens for pre-training. Similarly, ERNIE 1.0 and 2.0~\citep{sun2019ernie,sun2020ernie}, utilize named entity masking and phrase masking to encode the coarse-grained information into the models, while ERNIE-Gram~\citep{xiao2021ernie} uses explicit n-gram identities as predicted targets for the enhancement with coarse-grained information. Besides, \citet{joshi2020spanbert} propose the SpanBERT to mask text spans and train the span boundary objective. However, these methods mainly concentrate on fine-grained tokens. The coarse-grained information is only implicitly explored in the masked language modeling by designing the masking strategies and the coarse-grained representations are absent. 

Instead of designing the coarse-grained masking strategy on the fine-grained token sequences, several methods focus on improving the pre-training models with multi-grained tokenization. AMBERT \citep{zhang2021ambert} utilizes two encoders with shared parameters to process the fine-grained and coarse-grained token sequences. LICHEE \citep{guo2021lichee} proposes to merge the multi-grained tokenizations at the embedding level to incorporate multi-grained information of input. Recently, \citet{lai2021lattice} propose the Lattice-BERT, which introduces the lattice graph constructed from characters and words to explicitly explore the word representations in a multi-granularity way. 
However, these models are either computationally intensive or lack the integration of multi-grained information in the deep encoder layers, resulting in the limitations of usability and effectiveness. 

\begin{figure*}[th]
\centering
\includegraphics[width=2\columnwidth]{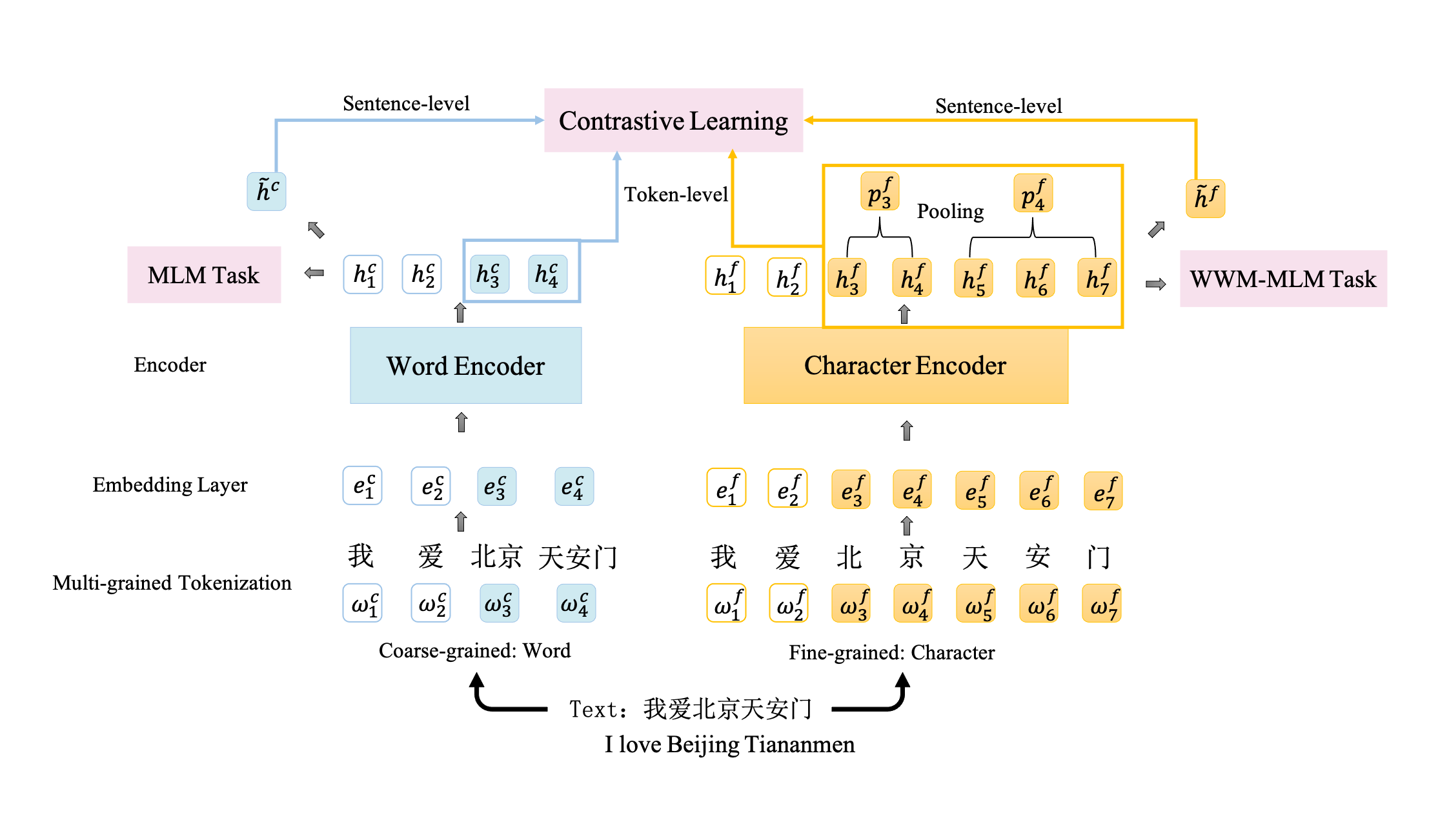}
\caption{An overview of \M. Fine-grained and coarse-grained representations are encoded by two encoders.Token-level and sentence-level contrastive learning are conducted together with the MLM and WWM-MLM tasks.}
\label{fig:model}
\end{figure*}

\paragraph{Contrastive Learning in Pre-trained Language Models}
As contrastive learning become popular in visual representation learning \citep{chen2020simple,he2020momentum, khosla2020supervised} and NLP tasks \citep{wu2020clear,meng2021coco, wang2021cline}, there have been several works exploring the effects of contrastive learning for pre-trained language models. CERT \citep{fang2020cert} adopts the framework of MOCO \citep{he2020momentum} 
and performs the sentence augmentations by back-translation. \citet{zhang2020unsupervised} propose the unsupervised sentence embedding model IS-BERT, 
increasing the mutual information between the global representations and the local context when training the model.
ConSERT \citep{yan2021consert} applies a variety of data augmentation techniques to generate various input views at the embedding level for contrastive learning.
Similarly, SimCSE \citep{gao2021simcse} utilizes dropout acts as data augmentation in sentence-level. The above methods conduct the contrastive learning to fine-tune the pre-trained language encoder. As for pre-training the language model, DeCLUTR~\citep{giorgi2021declutr} and CLEAR~\citep{wu2020clear} utilize the architecture of SimCLR~\citep{chen2020simple} to 
combine the contrastive learning objective with the masked language modeling. Compared to the above models, our \M~conducts the contrastive learning over word and character representations in pre-training and we have the flexibility to fine-tune it in specific downstream tasks.

\section{Methodology}

In this section, we present \M, the pre-trained language model based on contrastive learning over word and character representations. We first present the overall model architecture of \M, and then we introduce its details in the pre-training stage. Finally, we discuss the strategy of fine-tuning the model efficiently using only the fine-grained input.

\subsection{Model Architecture}
Figure \ref{fig:model} illustrates an overview of \M~ pre-training, where the contrastive learning framework is leveraged across multiple granularity information to enhance the representation ability of the model.

\M~takes the text sequences as input and performs multi-grained tokenization on the input to obtain the fine-grained and coarse-grained token sequences. It should be noted that the fine-grained and coarse-grained tokens share the same vocabulary, which aims at improving the alignment of embedding spaces between multi-grained tokens. In this paper, we treat the characters and words as fine-grained and coarse-grained tokens respectively. Formally, given the input text sequence $s$, we denote the fine-grained and coarse-grained token sequences by $\bm{s_f}=\{\omega_1^{f},\cdots,\omega_i^{f},\cdots,\omega_m^{f}\}$ and $\bm{s_c}=\{\omega_1^{c},\cdots,\omega_j^{c},\cdots,\omega_n^{c}\}$, where $m$ and $n$ are the lengths of two tokenized sequences.

Consistent with the shared vocabulary, \M~adopts the shared embedding layers to map the tokens $\omega_i^{f}$ and $\omega_j^{c}$ to the embedding representations $\bm{e_i^f}$ and $\bm{e_j^c}\in \mathcal{R}^d$ respectively, where $d$ is the dimension of the embedding. The fine-grained and coarse-grained embeddings are then passed to the two encoders to obtain the contextualized representations respectively. The encoders utilized in \M~can be any pre-trained language model and two encoders of fine-grained and coarse-grained have independent parameters. In this paper, we adopt Chinese BERT\cite{kenton2019bert} as the encoders.

Token-level and sentence-level contrastive learning are conducted over the fine-grained and coarse-grained contextualized representations from the above encoders, together with the traditional MLM task and WWM-MLM task. 

\subsection{Pre-Training}
\paragraph{Masked Language Model}In the pre-training stage, \M~adopts the MLM task at multi-grained levels. Specifically, we denote the masked fine-grained and coarse-grained token sequences as $\bm{\tilde{s_f}}$ and $\bm{\tilde{s_c}}$. The masked fine-grained and coarse-grained tokens are represented as $\bm{s_f^m}$ and $\bm{s_c^m}$ respectively. Then, the object of our MLM task at multi-grained levels is to optimize the following loss function:
\begin{equation}
\begin{split}
    \mathcal{L}_{mlm}= & -\sum_{\omega^m_f \in \bm{s_f^m}} \log P_{\theta}(\omega^m_f|\bm{\tilde{s_f}}) \\
    & -\sum_{\omega^m_c \in \bm{s_c^m}} \log P_{\theta}(\omega^m_c|\bm{\tilde{s_c}}),
\end{split}
\end{equation}
where $\theta$ denotes the model parameters.

We adopt the WWM strategy~\cite{cui2021pre} as the strategy of fine-grained token sequences and the conventional masking strategy introduced by BERT\cite{kenton2019bert} for the coarse-grained token sequences.

\paragraph{Contrastive Learning} To fully learn from the multi-grained information, we conduct contrastive learning between the fine-grained representations and their corresponding coarse-grained representations at both token-level and sentence-level. Formally, for each pair of multi-grained token sequences $\bm{s_f}$ and $\bm{s_c}$, we randomly choose some of the coarse-grained tokens $\bm{s_a} = \{ \omega_1^c, \cdots, \omega_i^c, \cdots, \omega_k^c\}\subset \bm{s_c}$ as anchors, where $k$ is the maximum number of anchors for each sequence. The strategy of selecting the anchors will be detailed in Section \ref{pre-training datasets}. 

Given the anchor $\omega_i^c$, which is composed of the fine-grained tokens $\omega_{b(i)}^{f}, \cdots, \omega_{e(i)}^{f}$ where $b(i)$ denotes the begin index of the anchor $\omega_i^c$ and $e(i)$ denotes the end index of the anchor $\omega_i^c$, we can obtain its coarse-grained representation $\bm{h_{i}^c}$ generated by the word encoder and its fine-grained representation $\bm{p_{i}^{f}}=\text{AVG}\left(\bm{h_{b(i)}^{f}} \cdots  \bm{h_{e(i)}^{f}}\right)$ generated by the character encoder, where $\text{AVG}(\cdot)$ means the average pooling.

Our motivation is to close the gap between the fine-grained representations and their corresponding coarse-grained representations while enlarge the gap between unrelated representations. Following the contrastive learning paradigm, it can be implemented by constructing positive and negative instance pairs. For the coarse-grained representation $\bm{h_{i}^c}$, we mark the fine-grained representations of the same anchor $\bm{p_{i}^{f}}$ as its positive instance and the fine-grained representations of the other anchors in the same mini-batch $\bm{p_{j}^{f}}$ as the negative instances. We further introduce the ``\texttt{[CLS]}'' embeddings of each sentence as the sentence-level representations, namely $\bm{\tilde{h}^c}$ for the coarse-grained representation and $\bm{\tilde{h}^f}$ for the fine-grained representation. Similar to the token-level, we treat the multi-grained representations $(\bm{\tilde{h}^c}, \bm{\tilde{h}^f})$ of the same sentence as the positive instance pair and the multi-grained representations of different sentences in a mini-batch as the negative instance pairs.

Following the contrastive objective in \citet{chen2020simple}, we utilize the normalized temperature-scaled cross-entropy loss (NT-Xent) for both the token-level and sentence-level representations. We optimize the symmetric cross-entropy loss in the pre-training. Specifically, the objective of contrastive learning in multi-grained token-level representations $\mathcal{L}_{tcl}$ is as follows:
\begin{gather}
    \mathcal{L}_{tcl}^{c}=-\frac{1}{N}\sum_{i=1}^N\log \frac{e^{{\rm sim}(\bm{h_{i}^c}, \bm{p_{i}^{f}})/\tau}}{\sum_j e^{{\rm sim}(\bm{h_{i}^c}, \bm{p_{j}^{f}})/\tau}}, \\
    \mathcal{L}_{tcl}^{f}=-\frac{1}{N}\sum_{i=1}^N\log \frac{e^{{\rm sim}(\bm{p_{i}^{f}},\bm{h_{i}^c})/\tau}}{\sum_j e^{{\rm sim}(\bm{p_{i}^{f}}, \bm{h_{j}^c})/\tau}}, \\
    \mathcal{L}_{tcl} = \frac{1}{2} (\mathcal{L}_{tcl}^{c}+\mathcal{L}_{tcl}^{f}),
\end{gather}
where N indicates the number of in-batch anchors, ${\rm sim}(\cdot)$ denotes the similarity function as we use the cosine similarity, and $\tau$ is a temperature hyper-parameter.
Similarly, we define the symmetric sentence-level contrastive loss $\mathcal{L}_{scl}$ with a mini-batch size M as:
\begin{gather}
    \mathcal{L}_{scl}^{c}=-\frac{1}{M}\sum_{i=1}^M\log \frac{e^{{\rm sim}(\bm{\tilde{h}_{i}^c}, \bm{\tilde{h}_{i}^{f}})/\tau}}{\sum_j e^{{\rm sim}(\bm{\tilde{h}_{i}^c}, \bm{\tilde{h}_{j}^{f}})/\tau}}, \\
    \mathcal{L}_{scl}^{f}=-\frac{1}{M}\sum_{i=1}^M\log \frac{e^{{\rm sim}(\bm{\tilde{h}_{i}^{f}},\bm{\tilde{h}_{i}^c})/\tau}}{\sum_j e^{{\rm sim}(\bm{\tilde{h}_{i}^{f}}, \bm{\tilde{h}_{j}^c})/\tau}}, \\
    \mathcal{L}_{scl} = \frac{1}{2} (\mathcal{L}_{scl}^{c}+\mathcal{L}_{scl}^{f}),
\end{gather}
Therefore, the final object of contrastive learning $\mathcal{L}_{con}$ is the sum of $\mathcal{L}_{tcl}$ and $\mathcal{L}_{scl}$.

\paragraph{Sentence Order Prediction} Apart from the MLM and contrastive learning tasks, we adopt the sentence order prediction (SOP) task~\citep{lan2019albert} to effectively model the relationship of sentence pairs and denote the training loss as $\mathcal{L}_{sop}$. Hence, the overall training loss of \M~in pre-training is the combination of three tasks:
\begin{equation}\label{eq:total loss}
    \mathcal{L} = \mathcal{L}_{mlm} + \lambda \mathcal{L}_{sop} + \mu \mathcal{L}_{con}
\end{equation}
where $\lambda$ and $\mu$ are the hyper-parameters of balancing three task objectives.

\subsection{Fine-Tuning}

Note that the usage of the character encoder of CLOWER is virtually the same as the fine-grained Chinese PLMs like BERT, thus we can directly substitute them with our character encoder without any modification while having the benefit of the coarse-grained information encoded in the fine-grained representations.

For the sentence-level downstream tasks, like single sentence classification and sentence pair classification, we conduct classification base on the contextualized sentence-level representation $\bm{\tilde{h}^f}$. As for the token-level tasks, such as question answering, fine-grained contextualized representations of each token are extracted and used for predictions.


\section{Experiments}\label{exp}
We conducted comprehensive experiments on various Chinese NLU tasks to examine the effectiveness of \M. 
In this section, we first introduce the details of pre-training and fine-tuning, including the datasets and experimental settings. 
Then, we present the overall results on different tasks and conduct an in-depth analysis. 
Ablation studies are also conducted to evaluate the impact of multi-level contrastive learning in our model.

\subsection{Pre-training Datasets}\label{pre-training datasets}
To the best of our knowledge, WuDaoCorpora~\citep{yuan2021wudaocorpora} is the largest open-source Chinese corpora for pre-training. 
We utilize the base version of WuDaoCorpora\footnote{\url{https://resource.wudaoai.cn/home}}, consisting of about $200$GB training data and $72$ billion Chinese characters in total. 
Following the settings of most Chinese PLMs, we consider the characters as the fine-grained tokens. 
We utilize Jieba\footnote{\url{https://github.com/fxsjy/jieba}} to perform the word segmentation on texts and the segmented words are treated as the coarse-grained tokens. 
There are $5, 466$ Chinese characters and $40, 014$ words in our vocabulary, together with other tokens like digits and some basic English tokens. 
We conduct the fine-grained and coarse-grained tokenizations based on the vocabulary and the words will be split to characters if they are not in the vocabulary. 
For contrastive learning, we select up to $k$ anchors whose lengths are between $2$ and $4$ from each sequence. 
Note that for semantic integrity, the words that have been masked either on coarse-grained sequences or their fine-grained characters will not be selected as anchors.

\subsection{Fine-tuning tasks}
\begin{table}[t]
    \centering
    \begin{adjustbox}{width=0.9\columnwidth,center}
    \scalebox{0.9}{
    \begin{tabular}{ccccc}
        \toprule
        \textbf{Dataset} & \textbf{MSL} & \textbf{BS} & \textbf{LR} & \textbf{Epoch} \\
        \midrule
        ChnSentiCorp & 256 & 32 & 3e-5 & 10  \\
        THUCNews & 512 & 16 & 3e-5 & 10 \\
        Tnews & 128 & 32 & 3e-5 & 10 \\
        \midrule
        Bq Corpus & 128 & 64 & 3e-5 & 10 \\
        Lcqmc & 128 & 64 & 3e-5 & 10 \\
        Ocnli & 128 & 32 & 3e-5 & 10 \\
        Xnli & 128 & 64 & 3e-5 & 10\\
        \midrule
        CMRC2018 & 512 & 8 & 3e-5 & 5 \\
        DRCD & 512 & 8 & 3e-5 & 5 \\
        \bottomrule
    \end{tabular}
    }
    \end{adjustbox}
    \caption{Hyper-parameters settings for 9 fine-tuning tasks. MSL: Maximum Sequence Length; BS: Batch Size; LR: Learning Rate.}
    \label{tab:datasets}
\end{table}

\begin{table*}[t]
    \centering
    \begin{tabular}{ccccccc}
    \toprule
        \multirow{2}{*}{\textbf{Model}} & \textbf{Tnews} & \multicolumn{2}{c}{\textbf{THUCNews}} & \multicolumn{2}{c}{\textbf{ChnSentiCorp}} & \multirow{2}{*}{\textbf{Average}} \\
         & \textbf{Dev} & \textbf{Dev} & \textbf{Test} & \textbf{Dev} & \textbf{Test} & \\
    \midrule
    BERT-wwm & 66.59 & 98.16 & 97.41 & 94.97 & 95.55 & 90.53 \\
    BERT-wwm-sop & 66.42 & 98.31 & 97.49 & 94.87 & 95.32 & 90.48 \\
    MM-BERT & 66.39 & 98.18 & 97.53 & 94.92 & 94.80 & 90.36 \\
    MM-BERT-sop & 66.27 & 98.16 & 97.45 & 94.62 & 95.65 & 90.43 \\
    MacBERT & 67.07 & 98.29 & 97.34 & 95.16 & 95.18 & 90.61  \\
    \midrule
    \M & \textbf{67.15} & \textbf{98.39} & \textbf{97.74} & \textbf{95.18} & \textbf{95.84} & \textbf{90.86} \\
    \bottomrule
    \end{tabular}
    \caption{Experimental results on single sentence classification tasks.}
    \label{tab:ssc_results}
\end{table*}


\begin{table*}[t]
    \centering
    \begin{tabular}{ccccccccc}
    \toprule
        \multirow{2}{*}{\textbf{Model}} & \textbf{Ocnli} &
        \multicolumn{2}{c}{\textbf{Lcqmc}} & \multicolumn{2}{c}{\textbf{Xnli}} &
        \multicolumn{2}{c}{\textbf{Bq}} & \multirow{2}{*}{\textbf{Average}} \\
         & \textbf{Dev} & \textbf{Dev} & \textbf{Test} & \textbf{Dev} & \textbf{Test} & \textbf{Dev} & \textbf{Test} & \\
    \midrule
    BERT-wwm & 74.87 & 89.37 & 86.93 & 79.50 & 78.89 & 85.39 & 84.37 & 82.76 \\
    BERT-wwm-sop & 75.73 & 89.75 & 87.30 & 79.74 & 78.45 & 85.73 & 84.81 & 83.07 \\
    MM-BERT & 75.34 & 89.55 & 87.08 & 79.56 & 78.66 & 85.37 & 84.51 & 82.87 \\
    MM-BERT-sop & 75.44 & 89.85 & 87.18 & 79.42 & 78.62  & 86.00 & 84.84 & 83.05 \\
    MacBERT & 75.90 & 89.58 & 86.59 & \textbf{80.54} & 79.10  & 85.71 & 84.95 & 83.20 \\
    \midrule
    \M  & \textbf{76.25} & \textbf{89.92} & \textbf{88.10} & 80.14 & \textbf{79.19} & \textbf{86.01} & \textbf{85.26} & \textbf{83.55} \\
    \bottomrule
    \end{tabular}
    \caption{Experimental results on sentence pair classification tasks.}
    \label{tab:spc_results}
\end{table*}

\begin{table}[t]
    \centering
    \begin{adjustbox}{width=1\columnwidth,center}
    \begin{tabular}{ccccccc}
    \toprule
        \multirow{3}{*}{\textbf{Model}} &
        \multicolumn{2}{c}{\textbf{CMRC2018}} & \multicolumn{4}{c}{\textbf{DRCD}} \\
         & \multicolumn{2}{c}{\textbf{Dev}} & \multicolumn{2}{c}{\textbf{Dev}} & \multicolumn{2}{c}{\textbf{Test}}  \\
         & \textbf{EM} & \textbf{F1} & \textbf{EM} & \textbf{F1} 
         & \textbf{EM} & \textbf{F1} \\
    \midrule
    
    BERT-wwm & 68.15 & 86.32 & 88.20 & 93.63 & 87.13 & 92.55 \\
    BERT-wwm-sop & 67.47 & 85.86 & 87.54 & 93.15 & 87.33 & 92.61 \\
    MM-BERT & 68.61 & 86.42 & 88.45 & 93.65 & 87.36 & 92.85 \\
    MM-BERT-sop & 67.57 & 86.18 & 88.30 & 93.50 & 87.18 & 92.76 \\
    MacBERT & 68.31 & 86.38 & \textbf{88.92} & \textbf{94.08} & \textbf{88.04} & \textbf{93.22} \\
    \midrule
    \M & \textbf{68.73} & \textbf{86.52} & 88.27 & 93.44 & 87.68 & 92.94 \\
    \bottomrule
    \end{tabular}
    \end{adjustbox}
    \caption{Experimental results on MRC tasks.}
    \label{tab:mrc_results}
\end{table}

To thoroughly examine the effectiveness of \M, an extensive set of experiments are performed on various Chinese NLU tasks, including three single sentence classification (SSC) tasks, four sentence pair classification (SPC) tasks and two machine reading comprehension(MRC) tasks.
Specifically, three SSC tasks are ChnSentiCorp~\citep{tan2008empirical}, THUCNews~\citep{li2007scalable} and Tnews~\citep{xu2020clue}; four SPC tasks include Bq Corpus~\citep{chen2018bq}, Lcqmc~\citep{liu2018lcqmc}, Ocnli~\citep{hu2020ocnli} and Xnli~\citep{conneau2018xnli}; two MRC tasks are CMRC2018~\citep{cui2019span} and DRCD~\citep{shao2018drcd}.

\subsection{Experiment Settings}

\subsubsection{Pre-training}
In pre-training of \M, 
we initiate both the character and word encoder with the Chinese BERT-base released by Google\footnote{\url{https://github.com/google-research/bert}} in order to reduce the total convergence time. 
Given a word not in the vocabulary, we initiate its embedding with the average pooling of the embeddings of the characters that make up the word.
For MLM tasks, as with the BERT, $15$\% of the tokens are masked randomly.
For token-level contrastive learning, the maximum number of anchors for each sequence is set as $20$ and the temperature is $0.05$. 
The hyper-parameters $\lambda$ and $\mu$ in Equation~\ref{eq:total loss} are both set as $1$. 
We set the maximum sequence length to $512$ throughout the pre-training and adopt the ADAM~\citep{kingma2014adam} optimizer with weight decay whose learning rate is $2e-5$. 
We train the model with a batch size of $960$ $(24 \times 40)$ for $300, 000$ steps. 
The pre-training is carried out on $40$ NVIDIA V100 GPUs. 
To improve efficiency, mixed precision training~\citep{micikevicius2017mixed} is adopted.

\begin{figure*}[t]
\centering
\includegraphics[width=1.8\columnwidth]{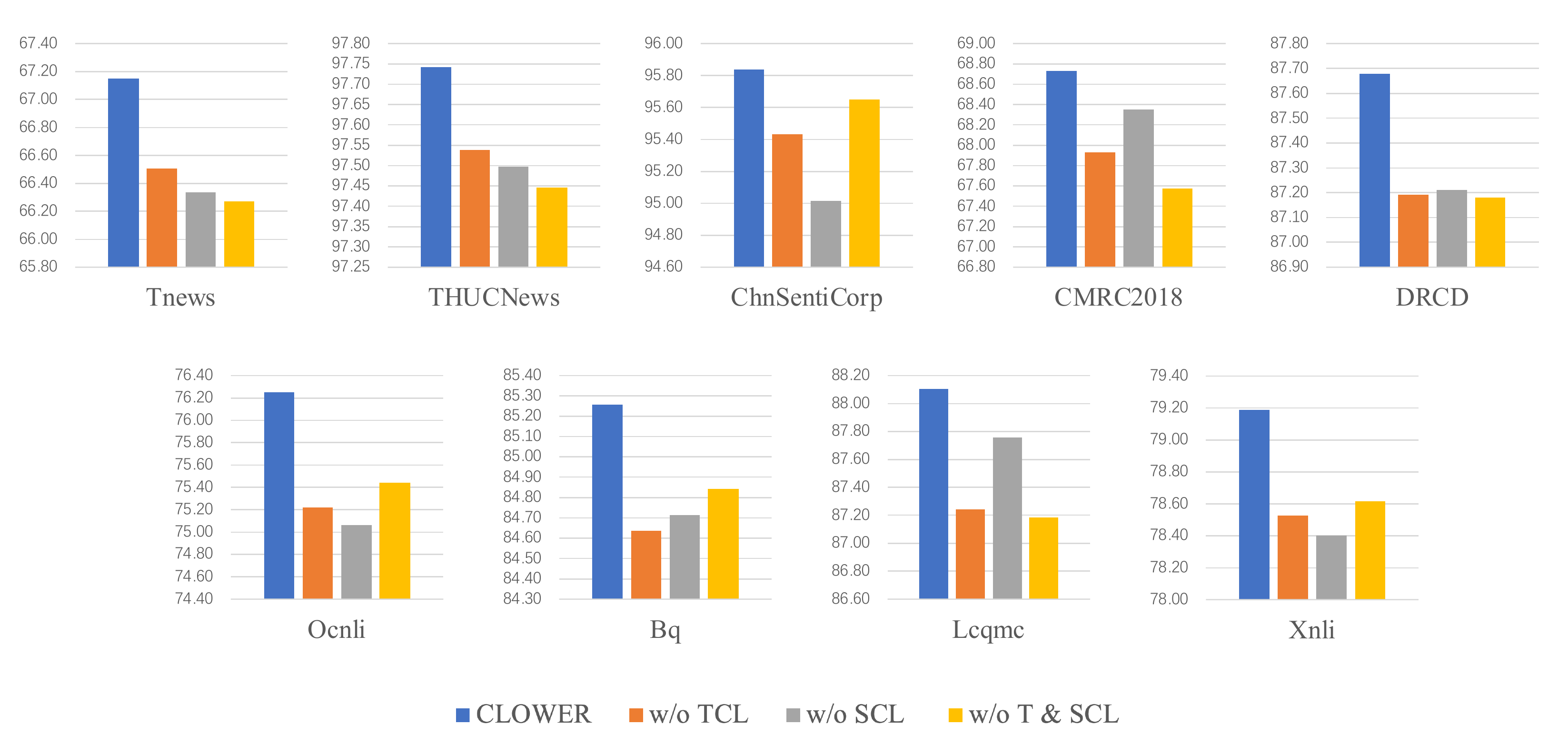}
\caption{Ablation Results. We report the accuracy for sentence classification tasks and EM for MRC tasks.}
\label{fig:abaltion}
\end{figure*}

\subsubsection{Fine-tuning}
To make a fair comparison, we adopt the same hyper-parameters for each fine-tuning task among different models.
The detailed parameter settings are shown in 
Table \ref{tab:datasets}. 
During fine-tuning, we encode each example using the fine-grained encoder (i.e., character encoder). 
For three SSP tasks and four SPC tasks, the ``\texttt{[CLS]}'' embedding is used to represent the sentence and the classification accuracies are reported.
For two MRC tasks, the token embeddings are used to extract the answer span from the sentence, and both exact match (EM) and F1-score are reported. 
For each task, we perform the experiments five runs with different random seeds and report the average performance to promise the results convincing. 
We report the results both on the development sets and test sets, except for Tnews, Ocnli and CMRC2018, whose test sets are not publicly available.
Since each article in the Tnews dataset consists of a title and several keywords, we associate the titles with keywords as the input sequences to perform the classification task. 
We fine-tune all the models for each downstream task on one NVIDIA V100 GPU.

\begin{table*}[t]
    \centering
    \begin{tabular}{ccccccccc}
    \toprule
        \textbf{Model} & \textbf{Tnews} & \textbf{THUC} & \textbf{Chn} & \textbf{Ocnli} & \textbf{Lcqmc} & \textbf{Xnli} & \textbf{Bq} & \textbf{Average}  \\
    \midrule
    
    \M & \textbf{67.15} & \textbf{97.74} & \textbf{95.84} & \textbf{76.25} & \textbf{88.10} & \textbf{79.19} & \textbf{85.26} & \textbf{84.22} \\
    
    \midrule
    
    w/o tcl & 66.51 & 97.54 & 95.43 & 75.22 & 87.24 & 78.53 & 84.64 & 83.59 \\
     
    w/o scl & 66.34 & 97.50 & 95.02 & 75.06 & 87.76 & 78.40 & 84.71 & 83.54 \\
    
    w/o tcl \& scl & 66.27 & 97.45 & 95.65 & 75.44 & 87.18 & 78.62 & 84.84 & 83.64 \\
    \bottomrule
    \end{tabular}
    \caption{Ablation results on SSC and SPC tasks. For Tnews and Ocnli, the results are on development sets and others are on test sets.}
    \label{tab:ablation_classification_results}
\end{table*}

\begin{table}[t]
    \centering
    \begin{adjustbox}{width=1\columnwidth,center}
    \begin{tabular}{ccccccc}
    \toprule
        \multirow{3}{*}{\textbf{Model}} &
        \multicolumn{2}{c}{\textbf{CMRC2018}} & \multicolumn{4}{c}{\textbf{DRCD}} \\
         & \multicolumn{2}{c}{\textbf{Dev}} & \multicolumn{2}{c}{\textbf{Dev}} & \multicolumn{2}{c}{\textbf{Test}}  \\
         & \textbf{EM} & \textbf{F1} & \textbf{EM} & \textbf{F1} 
         & \textbf{EM} & \textbf{F1} \\
    \midrule
    \M & \textbf{68.73} & \textbf{86.52} & 88.27 & 93.44 & \textbf{87.68} & \textbf{92.94}  \\
    \midrule
    w/o tcl & 67.93 & 86.25 & 88.01 & 93.34 & 87.19 & 92.69   \\
    
    w/o scl & 68.35 & 86.18 & 88.05 & 93.38 & 87.21 & 92.68 \\
    
    w/o tcl \& scl & 67.57 & 86.18 & \textbf{88.30} & \textbf{93.50} & 87.18 & 92.76 \\
    \bottomrule
    \end{tabular}
    \end{adjustbox}
    \caption{Ablation results on machine reading comprehension tasks.}
    \label{tab:ablation_mrc_results}
\end{table}

\subsection{Main Results}
Since most of the existing Chinese PLMs are trained with different corpus and setups, it is hard to conduct ideally fair comparisons. 
Therefore, we select the most representative Chinese PLM (i.e., Chinese BERT-base) as the baseline and achieve several pre-training models with different settings on the same corpus. 
More concretely, we implement the following four baselines: 
(1) \textbf{BERT-wwm}~\citep{cui2021pre}, a BERT-base model trained with the additional fine-grained WWM task, (2) \textbf{BERT-wwm-sop}, a BERT-base model trained with the addtional WWM and SOP tasks, (3) \textbf{MM-BERT}, a BERT-base model trained with the multi-grained MLM tasks, including a fine-grained WWM task and a coarse-grained MLM task, (4) \textbf{MM-BERT-sop}, Multi-grained MLM on a BERT-base model trained with the multi-grained MLM task and the SOP task. 
In addition, we also include MacBERT~\citep{cui2021pre} as a strong baseline, which is one of the state-of-the-art Chinese PLMs in literature.
MacBERT utilize the WWM as well as N-gram masking strategies together during pre-training. 
In terms of the masking implementation, MacBERT masks the word with a similar word rather than the [Mask] placeholder to improve the performance further.
The experimental results of MacBERT are achieved with the released model\footnote{\url{https://github.com/ymcui/MacBERT}} under the identical settings with the other baselines among all downstream tasks.

For three SSC tasks, the results are shown in Table~\ref{tab:ssc_results}. 
From the results, we can find that our \M~yields consistent improvements over all baselines on all three tasks (both on the development and test sets), which proves the effectiveness and advantages of our model. 
\M~ outperforms the $4$ baselines pre-trained with the identical data while different settings, which demonstrates the advantages of our multi-level contrastive learning approach.
In addition, \M~outperforms MacBERT by $0.25$ points on average and achieves a new state-of-the art on Chinese SSC tasks. 

As for the SPC tasks, fair comparisons are performed and the results are reported in Table~\ref{tab:spc_results}. 
From the results, we also observe that \M~also achieves consistent improvements over baselines on the four tasks.
In comparison to the four baselines pre-trained with the identical data, \M~outperforms the best one (i.e., MM-BERT-sop) by 0.33 points on average. 
In comparison to MacBERT, \M~achieves a performance gain of $0.33$ points on average.
\M~performs best on all datasets except Xnli Dev set. 

The above SSC and SPC tasks are all sequence-level tasks, to further examine the effectiveness of our model, we also perform comparisons on MRC tasks which are document-level span-extraction tasks. 
The resuls are depicted in Tabel~\ref{tab:mrc_results}. 
Specifically, for CMRC2018, \M~outperforms MacBERT by $0.40$ points and $0.14$ points on EM and F1 score respectively. 
As the EM score is a stricter measurement of machine reading comprehension, the improvements over MacBERT are considerable. 
While for DRCD, we find that the performance of \M~is not as competitive as the baselines. 
We conjecture that the reason may be the original dataset of DRCD is in Traditional Chinese whereas our pre-training corpus is in Simplified Chinese.
Although we convert the data to Simplified Chinese literally, there are some differences such as syntax and semantics yet, 
the performances of the pre-trained models may be affected inevitably.

\subsection{Ablation Study}

To further investigative the effects of contrastive learning over word and characters in \M, we conduct ablation study on the model variants without token-level or sentence-level contrastive learning tasks. 
Figure~\ref{fig:abaltion} shows the ablation results on sentences classification and machine reading comprehension tasks.
The detailed ablation results on $9$ downstream NLU tasks are reported in Table~\ref{tab:ablation_classification_results} and \ref{tab:ablation_mrc_results} respectively.

When removing the token-level contrastive learning task (w/o TCL) or sentence-level (w/o SCL) from \M, there is a distinct drop in the performance on sentence classification tasks (i.e., SSC and SPC).
Furthermore, when removing all the contrastive learning tasks, i.e., actually the MM-BERT-sop model, the performance is almost same as the w/o TCL or w/o SCL models. 
It indicates that only if the token-level contrastive learning task works jointly with the sentence-level contrastive learning task in pre-training, there will be a positive impact on the sentence-level downstream tasks. 
We conclude that it is vital for the model to encode the coarse-grained semantic information into the fine-grained sequences at token-level and sentence-level consistently when we apply it on sentence-level downstream tasks.

    
    

As for the MRC tasks, the EM score on CMRC2018 drops a lot when removing the token-level contrastive learning task, which demonstrates the effectiveness of token-level task on the extractive MRC task. 
While removing the sentence-level contrastive learning task, the EM metric of the model drops less than that without the token-level. 
Also, the performance of model without both contrastive learning tasks perform worst among these models on CMRC2018. 
The results on DRCD reveal the similar trend.

\section{Discussions}
\subsection{Flexibility}
Compared to other Chinese PLMs which utilize fine-grained and coarse-grained information, one notable advantage of \M~is the high flexibility of deployment.
In real-world scenarios, fine-grained PLMs are more popular due to its flexibility on processing inputs/outputs and low computational costs. 
Please recap that \M~ could be deemed as a fine-grained character encoder during inference, which is enhanced with the coarse-grained word encoder during pre-training.
In particular, if a production system already deploys a fine-grained Chinese PLM (e.g., the vanilla BERT), the fine-grained encoder of \M~can be adopted as a substitute without extra tailor cost seamlessly.
\M~also provides the coarse-grained encoder (i.e., word encoder) for scenarios where Chinese word sequences are designed as input.
The coarse-grained encoder of \M~has also been updated and acquired the knowledge from the large corpus during pre-training. 
We can make flexible choices according to downstream scenarios and conditions when utilizing \M.

\subsection{Multi-grained Information Modeling}

\begin{figure}[t]
\centering
\includegraphics[width=0.7\columnwidth]{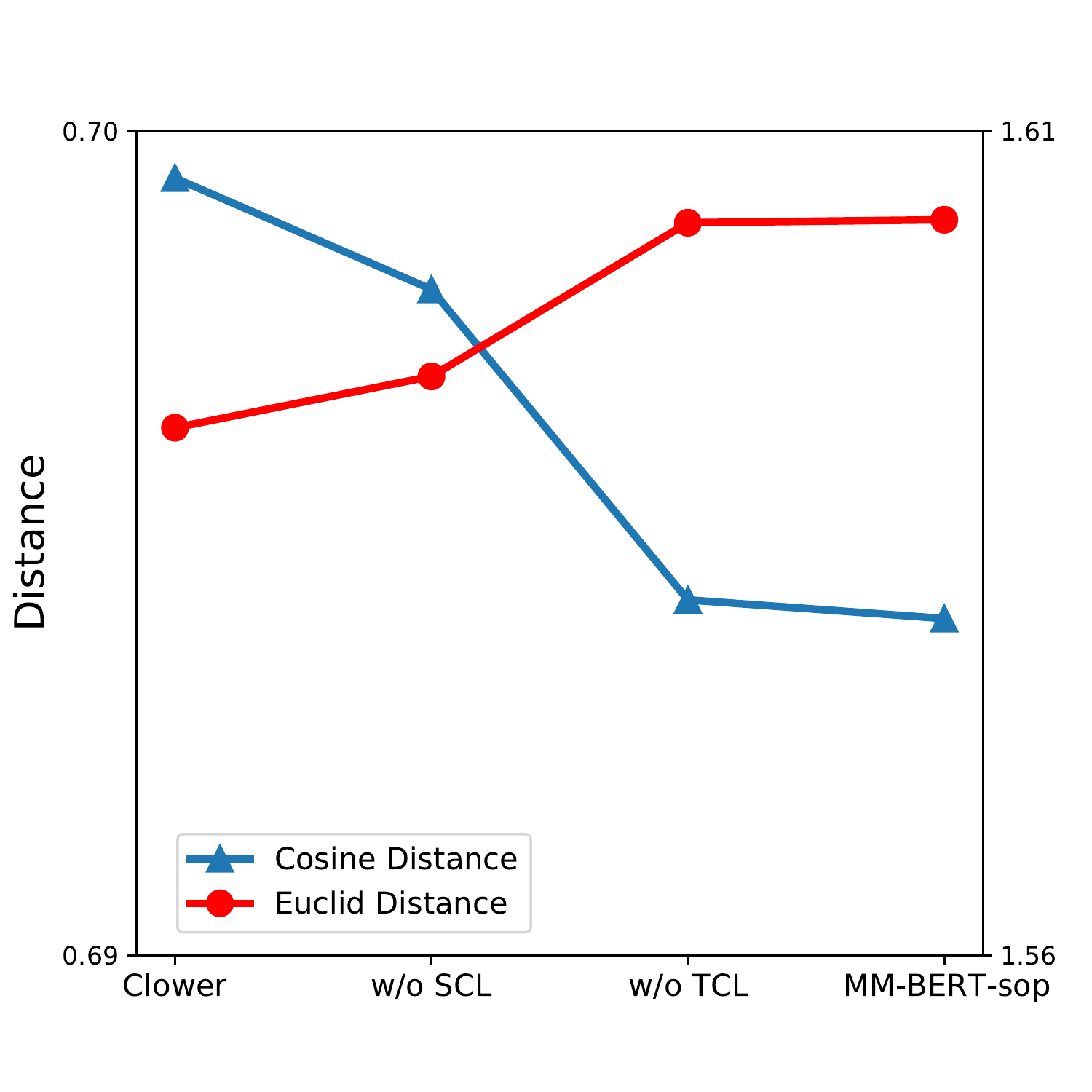}
\includegraphics[width=0.7\columnwidth]{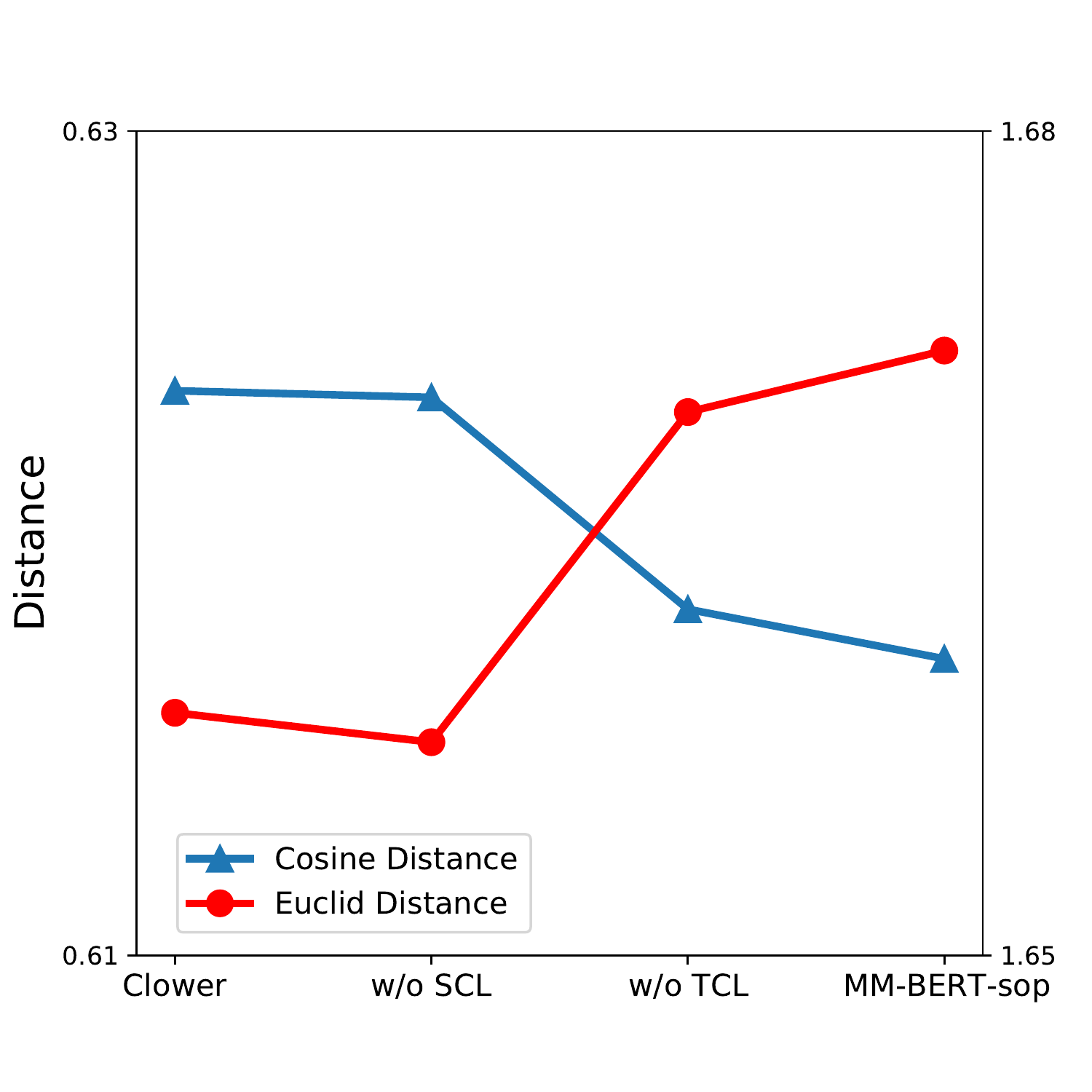}
\caption{Similarity Analysis of Embeddings. Top: the words with length 2; Bottom: the words with length longer than 2.}
\label{fig:sim_stat}
\end{figure}

Through the pre-training, \M~implements the multi-grained semantic information modeling by performing the contrastive learning over words to characters and thus implicitly encodes the coarse-grained semantic information into fine-grained tokens and vice versa.
To evaluate the character/word representations learned by the interactions, we adopt the measures of cosine similarity and Euclid distance as proxies.
We calculate the cosine similarity and Euclid distance between the embeddings of words and the mean embeddings of the characters that compose the words.
In our corpus, $72.1\%$ words are composed of two characters.
So we conduct the similarity analysis by split the words into two groups, two-character words and the other words composed at least three characters.
The similarities produced by four models
are shown in Figure~\ref{fig:sim_stat}. 
We can clearly see that the token-level contrastive learning task play an important role of bringing the word and character embeddings closer, as the similarity of \M~and w/o scl are higher than the other two models and so is the Euclid distance. 
According to the intuitive results, we corroborate that our model indeed achieves our motivation to encode the coarse-grained information into fine-grained tokens.

\section{Conclusion}
To fully leverage the information of characters and words in Chinese PLMs, we propose a novel PLM \M~based on contrastive learning over word and character representations jointly. 
Through the token-level and sentence-level contrastive learning in the pre-training stage, the model encodes the coarse-grained semantic information into fine-grained tokens. 
We can not only enhance the model with coarse-grained semantics but also enjoy the flexibility of fine-grained inputs/outputs. 
The flexibility promises that our model could be deployed conveniently in real scenarios, where certain PLMs like BERT have been established.
Comprehensive experiments on a variety of downstream natural language understanding tasks demonstrate the competitive performance of \M.
We also conduct a ablation study to evaluate the multi-grained contrastive learning mechanism in \M.


\section*{Acknowledgements}
This research is supported by  National Natural Science Foundation of China (Grant No. 6201101015), Beijing Academy of Artificial Intelligence (BAAI), Natural Science Foundation of Guangdong Province (Grant No. 2021A1515012640), the Basic Research Fund of Shenzhen City (Grant No. JCYJ20210324120012033 and JCYJ20190813165003837), National Key R\&D Program of China (No. 2021ZD0112905) and Overseas Cooperation Research Fund of Tsinghua Shenzhen International Graduate School (Grant No. HW2021008).

\clearpage
\normalem
\bibliography{anthology}
\bibliographystyle{acl_natbib}




\end{document}